\documentclass[10pt,twocolumn,letterpaper]{article}
\pdfoutput=1
\usepackage{iccv}
\usepackage{times}
\usepackage{epsfig}
\usepackage{graphicx}
\usepackage{amsmath}
\usepackage{amssymb}
\usepackage[ruled]{algorithm2e}
\usepackage{algorithmic}
\usepackage{caption}

\usepackage[breaklinks=true,bookmarks=false]{hyperref}

\iccvfinalcopy 

\def\iccvPaperID{10803} 

\begin{document}

\title{Photothermal-SR-Net: A Customized Deep Unfolding Neural Network for Photothermal Super Resolution Imaging}

\author{Samim Ahmadi, Jan Christian Hauffen, Mathias Ziegler\\
Bundesanstalt für Materialforschung und -prüfung (BAM)\\
Unter den Eichen 87, 12205 Berlin\\
{\tt\small \{samim.ahmadi,jan-christian.hauffen,mathias.ziegler\}@bam.de}
\and
Linh Kästner, Peter Jung\\
Technical University of Berlin\\
Straße des 17. Juni 135, 10623 Berlin \\
{\tt\small {doan.hl.kaestner@campus.tu-berlin.de,peter.jung@tu-berlin.de}
}}

\maketitle

\begin{figure*}[!h]
    \centering
    \includegraphics[width = 0.8 \textwidth]{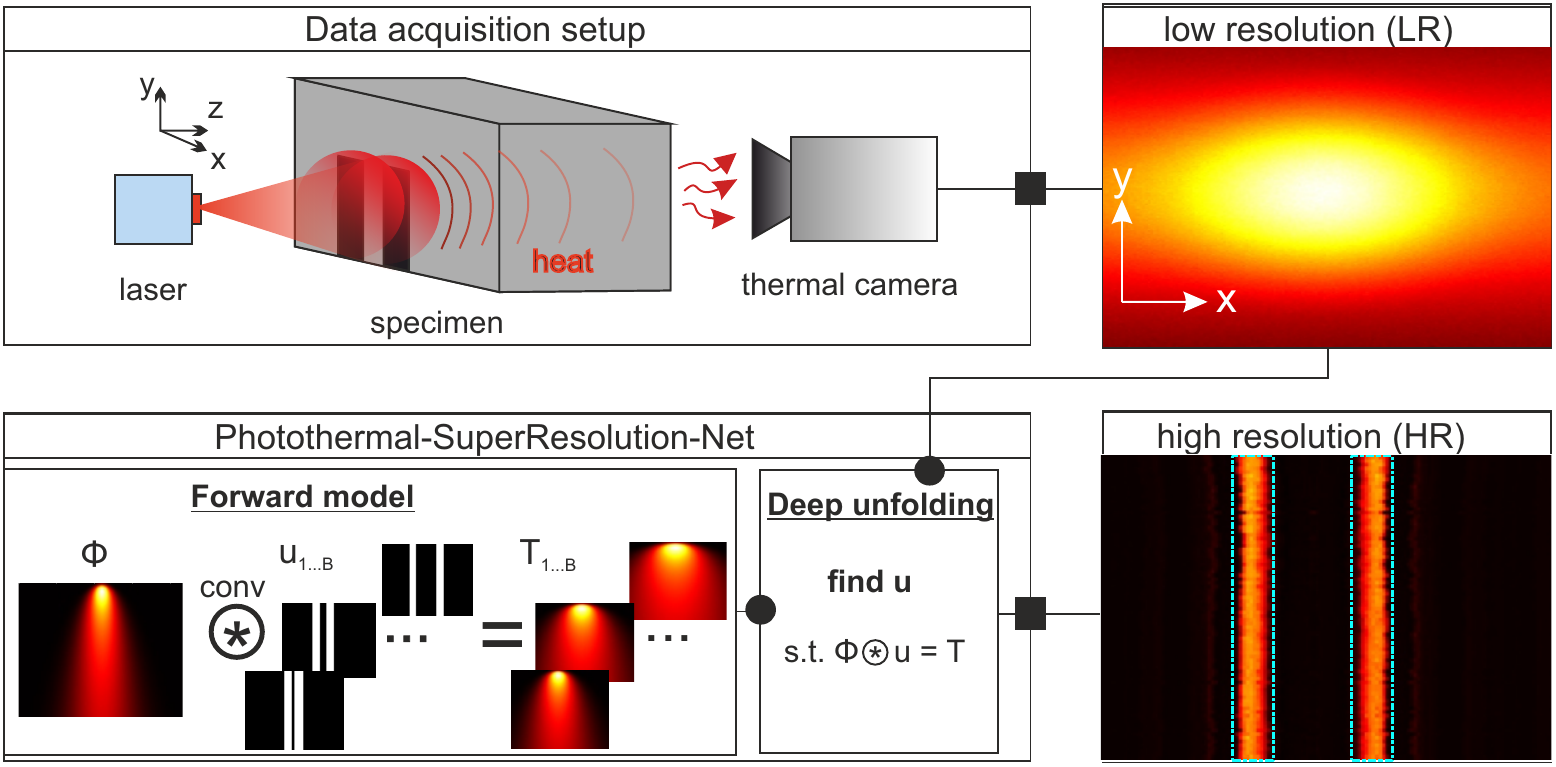}
    \caption{Photothermal-SuperResolution(SR)-Net used to convert low-resolution data from laser thermography experiment to high-resolution image. In the thermal images, yellow stands for hot and black for cold. The HR image shows the ground truth in cyan and the defect reconstruction with a hot colormap. 'conv' stands for convolution as well as $\ast$ denotes the convolution operator.}
    \label{fig:Einstiegsbild}
\end{figure*}

\begin{abstract}
   This paper presents deep unfolding neural networks to handle inverse problems in photothermal radiometry enabling super resolution (SR) imaging.
   Photothermal imaging is a well-known technique in active thermography for nondestructive inspection of defects in materials such as metals or composites. A grand challenge of active thermography is to overcome the spatial resolution limitation imposed by heat diffusion in order to accurately resolve each defect. The photothermal SR approach enables to extract high-frequency spatial components based on the deconvolution with the thermal point spread function. However, stable deconvolution can only be achieved by using the sparse structure of defect patterns, which often requires tedious, hand-crafted tuning of hyperparameters and results in computationally intensive algorithms. 
   On this account, Photothermal-SR-Net is proposed in this paper, which performs deconvolution by deep unfolding considering the underlying physics. This enables to super resolve 2D thermal images for nondestructive testing with a substantially improved convergence rate. 
   Since defects appear sparsely in materials, Photothermal-SR-Net applies trained block-sparsity thresholding to the acquired thermal images in each convolutional layer. 
   The performance of the proposed approach is evaluated and discussed using various deep unfolding and thresholding approaches applied to 2D thermal images. Subsequently, studies are conducted on how to increase the reconstruction quality and the computational performance of Photothermal-SR-Net is evaluated. 
   Thereby, it was found that the computing time for creating high-resolution images could be significantly reduced without decreasing the reconstruction quality by using pixel binning as a preprocessing step.
\end{abstract}

\section{Introduction}
"Super resolution" refers to the state of exceeding conventional resolution limits. For example, optical SR was used in optical microscopy to break the Abbe limit, which is based on the physics of electromagnetic radiation \cite{hell2007far}. In most imaging systems, however, the limits of spatial resolution are instead determined by the pixel resolution of the optical sensors used, \eg in cameras. In this case, geometric SR is employed to describe pure data processing algorithms that create high-resolution images from low-resolution images.

\noindent Photothermal SR is akin to optical SR in that it makes use of structured illumination \cite{willig2006sted} in experiments to make features more visible. In the case of photothermal imaging for nondestructive testing (NDT), narrow laser lines can be used to make these features, i.e. defects like cracks or voids in materials such as metals or fiber composites, more visible. In addition, it is common in optical SR to apply image processing algorithms in post-processing that make use of deconvolution with the optical point spread function, which leads to an even higher local resolution \cite{zhang2019super}. Photothermal SR, similar to optical SR, utilizes the fundamental solution of the underlying fundamental equation to achieve high spatial resolution. In case of photothermal SR, this equation is the heat diffusion equation and its solution is called thermal point spread function \cite{burgholzer2017super}. Therefore, the physical diffusion limit present in this case can be interpreted analogously to the diffraction limit in the case of optics \cite{burgholzer2015thermodynamic}. 

\noindent Higher spatial resolution, obtained by applying photothermal SR, enables improved quality assurance in the production industry, like automotive industry, medical technology, and additive manufacturing for 3D printing processes. Apart from industry, there are possible applications for photothermal SR in medicine for early detection of small tumors and initiation of preventive measures and treatments. Hence, SR imaging is an emerging and increasingly used technique, and the required hardware becomes more advanced every year. However, the processing time scales with the amount of image data, which is particularly high for high-resolution imaging.

\noindent SR imaging, therefore, still benefits from fast image processing algorithms enabling the generation of high-resolution images. Photothermal-SR-Net considers the underlying physics in photothermal nondestructive testing and quickly generates (within a second) high-resolution 2D thermal images for improved quality inspection of materials. 
Considering factors like defect sparsity and customized convolutional layers based on the thermal point spread function, Photothermal-SR-Net restricts the solution space and thus enables high rates of convergence. An illustration of a possible application for Photothermal-SR-Net is shown in Fig. \ref{fig:Einstiegsbild}. 

\noindent Therefore, Photothermal-SR-Net could be used for in-situ inspection in photothermal imaging, for example, to visualize defects in high resolution during the 3D printing process in additive manufacturing. The main contributions of this paper can be summarized as follows:
\begin{itemize}
    \item Applications of different customized deep unfolding neural networks to solve the underlying thermal inverse problem and reconstruct high-resolution spatial 2D thermal images for nondestructive testing with a high computational performance.
    \item Studies of the influence of tied and untied learning with weights based on the thermal point spread function as well as different implementations of activation functions using sparsity regularization like Block Fast Iterative Shrinkage Thresholding Algorithm (Block-FISTA) and Block Fast Elastic Net (Block-FENet) with or without rectified linear unit (ReLU) after applying gradient descent.
    \item Evaluations of 1D spatial pixel binning to obtain high defect reconstruction qualities with smaller computing times. 
\end{itemize}

\section{Related Work} 
In this section, work related to Photothermal-SR-Net is described and referenced. \newline
\textbf{Photothermal SR for NDT.} \quad Photothermal imaging is used in a wide field of applications ranging from microscopy \cite{nedosekin2014super,li2017super}, forensics \cite{cui2015highly,furstenberg2008stand,furstenberg2012chemical} to material research \cite{boyer2002photothermal,zharov2005photothermal}.
Due to the blurring effect induced by heat diffusion, thermal images often needs to be resolved in high quality.
Therefore, various approaches were introduced in literature. Graupy et al. \cite{grauby1999high} introduced a charge coupled device to generate high frequencies and accomplish high resolution of thermal images. Recent work employed thermal imaging to detect nano-objects using large scattering cross sections \cite{shi2020photothermal}. Super resolution is often necessary for more specific applications like quality assessment or microbiology. Nedoskin et al. \cite{nedosekin2014super} employ nonlinear photothermal microscopy for super resolution. Sunian et al. \cite{sunian2020resolution} enhanced the resolution by high order correlations. 
Photothermal SR for nondestructive testing was first presented in \cite{burgholzer2017super}, which was closely related to the demonstrated SR technique in photoacoustics \cite{murray2017super}. Laser sources have also been used in the past to realize photothermal super resolution \cite{burgholzer2020blind}. Like in optical SR, photothermal SR makes use of structured illumination in the experiment (c.f. data acquisition setup in Fig. \ref{fig:Einstiegsbild}) as well as of a deconvolution algorithm using the thermographic point spread function (PSF) in postprocessing. The deconvolution relies on the forward model as shown in Fig. \ref{fig:Einstiegsbild} (see the Photothermal-SR-Net box). Thereby, $\Phi$ represents the thermal PSF, $u_{1 \dots B}$ the ground truth/ defect patterns and $T_{1 \dots B}$ the measured temperature with the thermal camera, where $B$ denotes the number of batches used in the training. The problem that is solved by deconvolution is shown in Fig. \ref{fig:Einstiegsbild} in the "Deep unfolding" box. \newline
\textbf{Blind structured illumination.} \quad The aforementioned publications find $u$ by using a least squares term and block-sparsity regularization with an $\ell_{2,1}$-norm. The reason for introducing block-sparsity with $\ell_{2,1}$ instead of simple sparsity with $\ell_1$ is that blind illumination is assumed, where one does not know the exact position of laser illumination. This assumption simplifies the usage of photothermal SR since the user does not have to know the exact position and the model given by the thermal PSF $\Phi$ is easier to determine. Also, the exact position of illumination can be influenced by positional noise if the laser is, for instance, held by a robot so that using $\ell_1$ with an estimation of the illuminated position could result in worse performance than using $\ell_{2,1}$ without considering the illuminated position \cite{mudry2012structured,burgholzer2017super}. \newline
\textbf{Block-sparsity regularization for defect reconstruction.} \quad 
Block-sparsity regularization is a technique to solve  underdetermined equation systems by using the known structure of the target, i.e. it is known, that only few elements are nonzero and these elements occur in coherent blocks \cite{eldar2008blocksparsity}. This finds widespread application \cite{wang2019reweighted}.
\noindent Cai et al. \cite{cai2018block} used sparsity regularization to reconstruct images from tomography. Xie et al. \cite{xie2016multispectral} propose a tensor-based denoising approach to improve image quality in multispectrum. A suitable block-sparsity optimization algorithm called Block Fast Iterative Shrinkage Thresholding Algorithm (Block-FISTA) was presented in \cite{murray2017super,burgholzer2017super} considering block-sparsity regularization. Ahmadi et al. \cite{ahmadi2020photothermal} recognized that the preprocessing steps before applying the Block-FISTA approach are also important for the final results. Data reduction as preprocessing is required because the datasets generated by photothermal SR measurements are large and the deconvolution performance is not high enough to handle these large datasets. The influence of the chosen experimental parameters on the data acquisition, as well as the impact of the chosen regularization parameters on the optimization algorithms, such as in Block-FISTA or Block Elastic Net (Block-ENet), were presented in \cite{ahmadi2020super}. The difference between the ENet approach and the ISTA approach is that the former uses an additional Tikhonov regularization term while ISTA only relies on sparsity regularization \cite{bach2012optimization,parikh2014proximal}. Moreover, the difference between Fast ISTA/ Fast ENet and ISTA/ENet is an additional step in updating the current iteration to reach a faster convergence \cite{beck2009FISTA}. Also, block-sparsity regularization has been applied in combination with virtual wave image processing to photothermal SR data for multi-dimensional defect reconstruction \cite{ahmadi2021multi}.  \newline
\textbf{Deep unfolding for photothermal SR.} \quad Deep unfolding is an emerging technique used in various areas especially in communication \cite{balatsoukas2019deep,wisdom2016deep} and signal processing \cite{hershey2014deep,hu2020iterative,lin2020unsupervised,cohen2019deep}. 
In recent years, deep unfolding networks emerged as a superior method for use cases such as image reconstruction, super resolution and denoising \cite{lucas2018using,ning2020accurate,liu2021sgd,ma2019deep,kim2020element}.
Ma et al. \cite{ma2019deep} proposed a deep unfolding network using the $\ell_1$-minimization to reconstruct images from a small amount of measurements. Zhang et al. \cite{zhang2020deep} proposed an end-to-end unfolding network to accomplish super resolution in noisy images. 

\noindent In \cite{zhang2020amp}, the researchers presented AMP-Net to denoise compressed images using deep unfolding networks. The researchers concluded an enhanced reconstruction accuracy by utilizing de-blocking modules to eliminate artefacts. 
Other works employed deep unfolding networks for guided super resolution of multispectral, near-infrared or ultrasound images \cite{bertocchi2020deep,cohen2019deep,marivani2020multimodal}. 

\noindent The main benefit of Photothermal-SR-Net is the higher rate of convergence. Deep unfolding in Photothermal-SR-Net enables to circumvent manually or empirically chosen regularization parameters. A first approach to implement a Photothermal-SR-Net was already shown in \cite{ahmadi2020learned}. However, only one optimization method was presented, called Block-ISTA, which has been combined with deep unfolding. Further, only spatial 1D thermal signals have been processed, whereas in this paper the application of Photothermal-SR-Net to 2D thermal images is proposed. Moreover, the performance of Photothermal-SR-Net with different advanced optimization routines such as Block-FISTA and Block Fast Elastic Net (Block-FENet) is compared.

\section{Methodology} 
\subsection{Experiment and Data Acquisition}
\begin{figure}[h]
    \centering
    \includegraphics[width = 0.37 \textwidth]{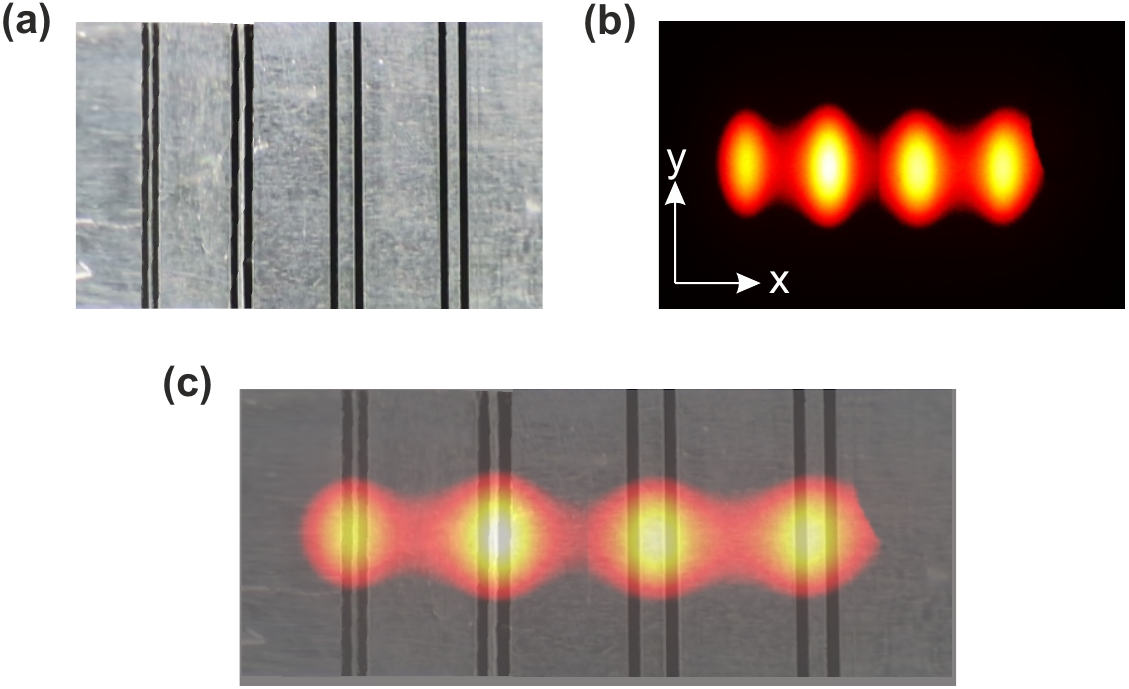}
    \caption{(a): Sample made of $3\,$mm thick blackened steel, aluminum foil is glued over the sample surface and slits were inserted. The distances between the slits within a slit pair are for each slit pair from left to right, $0.5$, $1$, $2$, $1.3\,$mm, respectively. The distance from one slit pair to another is $10\,$mm. The reflectivity of aluminium foil is very high for the laser wavelength that has been used ($980\,$nm), whereas the reflectivity of the blackened steel is very small so that most of the heat is only generated in the slits; (b): sum of all measurements from photothermal SR experiment in the two spatial dimensions x and y, yellow is hot and black is cold; (c): The overlap of (a) and (b) is shown to visualize which region has been heated.}
    \label{fig:specimen}
\end{figure}

The data acquisition setup is illustrated in Fig. \ref{fig:Einstiegsbild}, whereas the real defect geometries of the specimen are shown in Fig. \ref{fig:specimen} (a). Fig. \ref{fig:specimen} (b) shows the sum of all measured thermal images after step scanning the specimen with structured pulsed laser illumination, which will hereinafter be declared as low-resolution data. More precisely, the same experimental strategy as explained in \cite{ahmadi2020photothermal} has been used. Hence, each slit pair of the specimen has been illuminated with $30$ laser pulses, each at different positions. As four slit pairs are investigated, $120$ structured illumination measurements have been performed. A cooling/ waiting time of $20\,$seconds between the measurements was ensured so that each measurement is comparable. While \cite{ahmadi2020photothermal} only refers to spatial 1D data (x-dimension), this work demonstrates the acquisition and processing of spatial 2D data (x- and y- dimension).

\subsection{Data Processing Using Photothermal-SR-Net}
\begin{figure*}[h]
    \centering
    \includegraphics[width = 0.95\textwidth]{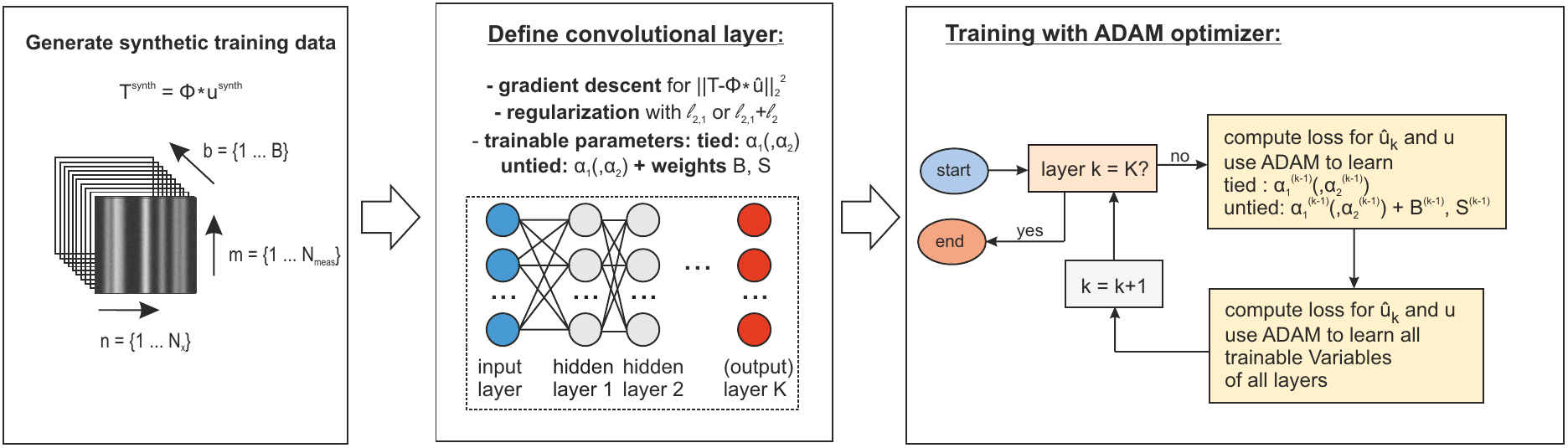}
    \caption{Flow chart to the steps in Photothermal-SR-Net: First the synthetic training data is generated. Afterward, the convolutional layers are defined where each layer consists of the gradient descent with subsequent regularization based on block-sparsity regularization for LBFISTA or based on block-sparsity and Tikhonov regularization for LBFENet. The next frame shows the implementation of the training by using the ADAM optimizer for each layer k.}
    \label{fig:flow_chart}
\end{figure*}

\textbf{Forward model.} \quad The acquired data, designated in the following as the measured temperature $T$, can be described by the following forward model
\begin{align} \label{eq:forward}
    T = \Phi \ast_{x,y,t} u,
\end{align}
whereby $\ast_{x,y,t}$ denotes the convolution in the two spatial dimensions $x$ and $y$ and the temporal dimension $t$. $\Phi \in \mathbb{R}^{N_x \times N_y \times N_t}$ denotes the discrete equivalent of the thermal point spread function $\phi: \mathbb{R}^3 \rightarrow \mathbb{R}_+$, which can be described by the Green's function \cite{cole2010heat}, with 
\begin{equation} \label{eq:phi}
    \phi(x,y,t) = \frac{2}{4 \pi \alpha \rho c_p} e^{-\frac{(x^2+y^2)}{4 \alpha t}}\cdot \sum_{p=-\infty}^{p = \infty} R^{2(p-1)}e^{-\frac{(2pL+z)^2}{4 \alpha t}}
\end{equation} for $t>0$. Thereby, $\rho$ stands for the mass density of the material (here structural steel S235JR), $c_p$ for the specific heat, $\alpha$ for the thermal diffusivity, $p$ for the number of reflections of the thermal wave in the material, $R$ for the thermal reflectance at the boundaries of the sample (material/ air), $L$ for the thickness of the sample and $z$ for a spatial dimension as shown in the Data acquisition setup in Fig. \ref{fig:Einstiegsbild}. Note that $z$ refers to the position at which the camera observes the surface of the specimen so that $z \in \{0,L\}$ with $z=0$ in reflection configuration and $z=L$ in transmission configuration. In this paper, we use transmission configuration, i.e. $z=L$. The following values have been used: $\rho = 7800\,\text{kg/m}^3$, $c_p = 440\,\text{J/kg/K}$, $\alpha = 1.6\cdot 10^{-5}\,\text{m}^2$/s, $R=1$, $L = 3\,$mm, $p=\{1,\,\dots,\,5\}$. The heat flux density $u \in \mathbb{R}^{N_x \times N_y \times N_t}$ is determined by $u = I \circ a$, where $I \in \mathbb{R}^{N_x \times N_y \times N_t}$ describes the spatial and temporal distribution of the laser illumination and $a \in \mathbb{R}^{N_x \times N_y}$ denotes the absorptance in space. More precisely, $I$ can be described by $I = I_{x,y} \otimes I_t$, where $\otimes$ denotes a tensor product and  $I_{x,y}\in \mathbb{R}^{N_x \times N_y}$ and $I_t \in \mathbb{R}^{N_t}$ stand for the spatial and temporal distribution of the illumination, respectively. To avoid a dimension clash in the Hadamard product $I \circ a$, the matrix $a$ can be repeated $N_t$ times. The forward model in eq. \eqref{eq:forward} does not consider noise, since camera noise can be removed by calculating the difference between the whole measured film sequence and the first frame of the film sequence before the laser measurement started. Hence, $T \in \mathbb{R}^{N_x \times N_y \times N_t}$ describes temperature differences and not absolute temperature values. Furthermore, this forward model does not consider boundary conditions like convection at the sample surface, assumes an isotropic material and no internal heat sources. Photothermal SR relies on multiple measurements using structured laser illumination such that only a small region of the sample surface is illuminated by one measurement $m$. To scan the whole sample surface, $N_{\text{meas}}$ measurements are necessary. In the present work, the experimental data rely on $N_{\text{meas}} = 120$ measurements, $30$ measurements per slit pair. Hence, equation \eqref{eq:forward} can be modified to:
\begin{align}
    T^m = \Phi \ast_{x,y,t} u^m
\end{align}
with $m = \{1,\,\dots,\,N_{\text{meas}}\}$. $\Phi$ does not change since the thermal point spread function remains the same for each measurement, as the same type of illumination is used. The only difference is given by the utilized illumination pattern, which changes $I$ to $I^m$ and consequently, $u$ to $u^m \in \mathbb{R}^{N_x \times N_y \times N_t}$, and $T$ to $T^m \in \mathbb{R}^{N_x \times N_y \times N_t}$.
\newline
\textbf{Blind structured illumination.} \quad
In this work, the illuminated position has not been used which means that $I_{x,y}$ cannot be extracted from $I$. On the one hand, this results in a huge solution space for $u^m$ solving the underlying inverse problem. On the other hand, the user does not have to know the exact position, which makes the application of the proposed approach more practical. To reduce the solution space for $u^m$, one can separate the illumination duration $I_t$ from $I^m$ and, therefore, from $u^m$, which yields:
\begin{align}
\begin{split}
    T^m &= (\Phi \ast_t I_t) \ast_{x,y} u_{x,y}^m\\
        &= \Phi_t \ast_{x,y} u_{x,y}^m,
\end{split}
\end{align}
where $\ast_{x,y}$ denotes the convolution in both spatial dimensions, $\Phi_t  \in \mathbb{R}^{N_x \times N_y \times N_t}$ and $u_{x,y}^m \in \mathbb{R}^{N_x \times N_y}$ denote the discrete equivalent of the thermal PSF, taking into account the pulse length in time-domain, and the heat flux density not considering the illumination duration, respectively.
Thus, a spatial deconvolution with $\Phi_t$ yields $u^m_{x,y}$.
\newline
\textbf{Training With Deep Unfolding.} \quad The training in Photothermal-SR-Net has been performed in x-dimension, since laser line excitation has been used and the pattern does not significantly change over the y-dimension. In addition, the training time would have been highly increased if the y-dimension had been considered. Thus, the training is performed based on unfolding the following forward model
\begin{align}
    T_{x,\,b = 1 \dots B}^{m,\,\text{synth}} = \Phi_{x,t} \ast_{x} u^{m,\,\text{synth}}_{x,\,b = 1 \dots B},
\end{align}
whereby $b$ denotes the batch number, since the pattern of $u$ varies from batch to batch. 'synth' stands for synthetic as the training was performed with synthetic data and not with experimental data. The shape of $u_{x,b}^m \in \mathbb{R}^{N_x}$ has been determined choosing the following quantities: sparsity, slit width, absorptance, SNR, laser line width.
$\Phi_x$ has been determined according to the underlying Green's function as described in eq. \eqref{eq:phi} without considering the variable $y$ and $\Phi_{x,t} = \Phi_x \ast_t I_t$, where $I_t$ describes the discrete equivalent of a rectangular function, which is one if the laser radiates and is zero otherwise. A spatial convolution in x-dimension ($\ast_x$) allows to generate synthetic temperature measurements $T_{x,\,b}^{m,\,\text{synth}} \in \mathbb{R}^{N_x}$. The deep unfolding is implemented as described in \cite{ahmadi2020learned}, while the thresholding algorithm used in the proposed Photothermal-SR-Net is different. In \cite{ahmadi2020learned}, the Learned Block Iterative Shrinkage Thresholding Algorithm (LBISTA) was used, whereas the present paper additionally proposes the use of Learned Block Fast Iterative Shrinkage Thresholding Algorithm (LBFISTA), Learned Block Fast Elastic Net (LBFENet) as well as Learned Block Elastic Net (LBENet) algorithm to compute an estimate $\hat{u} \approx u$. For the sake of simplicity, the following is written in the algorithm description: $T$ instead of $T_{x,\,b}^{m,\,\text{synth}}$, $\Phi$ instead of $\Phi_{x,t}$, $u$ instead of $u_{x,\,b}^{m,\,\text{synth}}$ and $\ast$ instead of $\ast_x$.
\begin{algorithm} \label{alg:lbfista}
    \SetKwInOut{Input}{Input}
    \SetKwInOut{Output}{Output}
    \Input{$T$}
    \Output{$\hat{u}$}
    $\hat{u}^{(0)}= B \ast T$\\
    $\hat{u}^{(1)}=\eta_{\left(\alpha_1^{(0)}, 0\right)}\left(\hat{u}^{(0)}\right)$\\
    $t_1 = \frac{1+\sqrt{5}}{2} $ \\
    $z^{(1)}=\hat{u}^{(0)}$\\
    \For{$k=2,\dots, K$}
   {
   $\hat{u}^{(k)}=\eta_{\left(\alpha_1^{(k-1)}, 0\right)}\left(S z^{(k-1)}+B \ast T \right)$\\
   $t_{(k)}=\frac{1+\sqrt{1+4t_{(k-1)}^2}}{2}$
   $z^{(k)}= \hat{u}^{(k)}+\frac{t_{k-1}-1}{t_{k}}\left(\hat{u}^{(k)}- \hat{u}^{(k-1)} \right)$
   }
    \caption{LBFISTA, layer definition: tied learning}
\end{algorithm}
$B = 2\gamma \Phi$ and $S = E-B \ast \Phi$, whereby $\gamma \in \mathbb{R}$ stands for the step size, and $E$ stands for a unit vector with ${E = [1,\, 0,\, \dots,\, 0]^T \in \mathbb{R}^{N_x}}$. The LBFENet algorithm differs from LBFISTA in that it uses $\alpha_2$, while LBFISTA sets $\alpha_2 = 0$ with 
\begin{align}
\begin{split}
    \eta&_{(\alpha_1, \alpha_2)}(u_x^m)[n] \\&=\max\left\lbrace0,1-\frac{\alpha_1}{\sqrt{\sum_{m=1}^{N_{meas}} |u_x^m[n]|^2}}\right\rbrace \frac{u_x^m[n]}{1+\alpha_2}  . \label{blocksoft} 
\end{split}
\end{align}
LBENet is implemented as LBFENet without considering the updating step in the second line of the for-loop with $t_{(k)}$. The implementation of the layer with tied and untied learning and the training itself has been implemented as described in \cite{ahmadi2020learned}, which was inspired by \cite{borgerding2017amp}. However, several training parameters have been chosen differently since one refinement has been used ($f_m = 0.5$) and the initial values for $\alpha_1$ and $\alpha_2$ have been chosen as $0.1$. The maximum number of iterations within the ADAM optimizer has been set to $5000$ in both - tied and untied learning. 

\noindent The whole Photothermal-SR-Net can be summarized by the flow chart shown in Fig. \ref{fig:flow_chart}. The training took approximately six hours using one GPU (Quadro RTX 5000). In this work, the generation of high-resolution 2D thermal images is realized by applying the Photothermal-SR-Net to each pixel row in the y-dimension individually.

\section{Results and Discussion} 
\textbf{Defect reconstruction with Photothermal-SR-Net.}
\begin{figure*}[h]
    \centering
    \includegraphics[width = 0.9\textwidth]{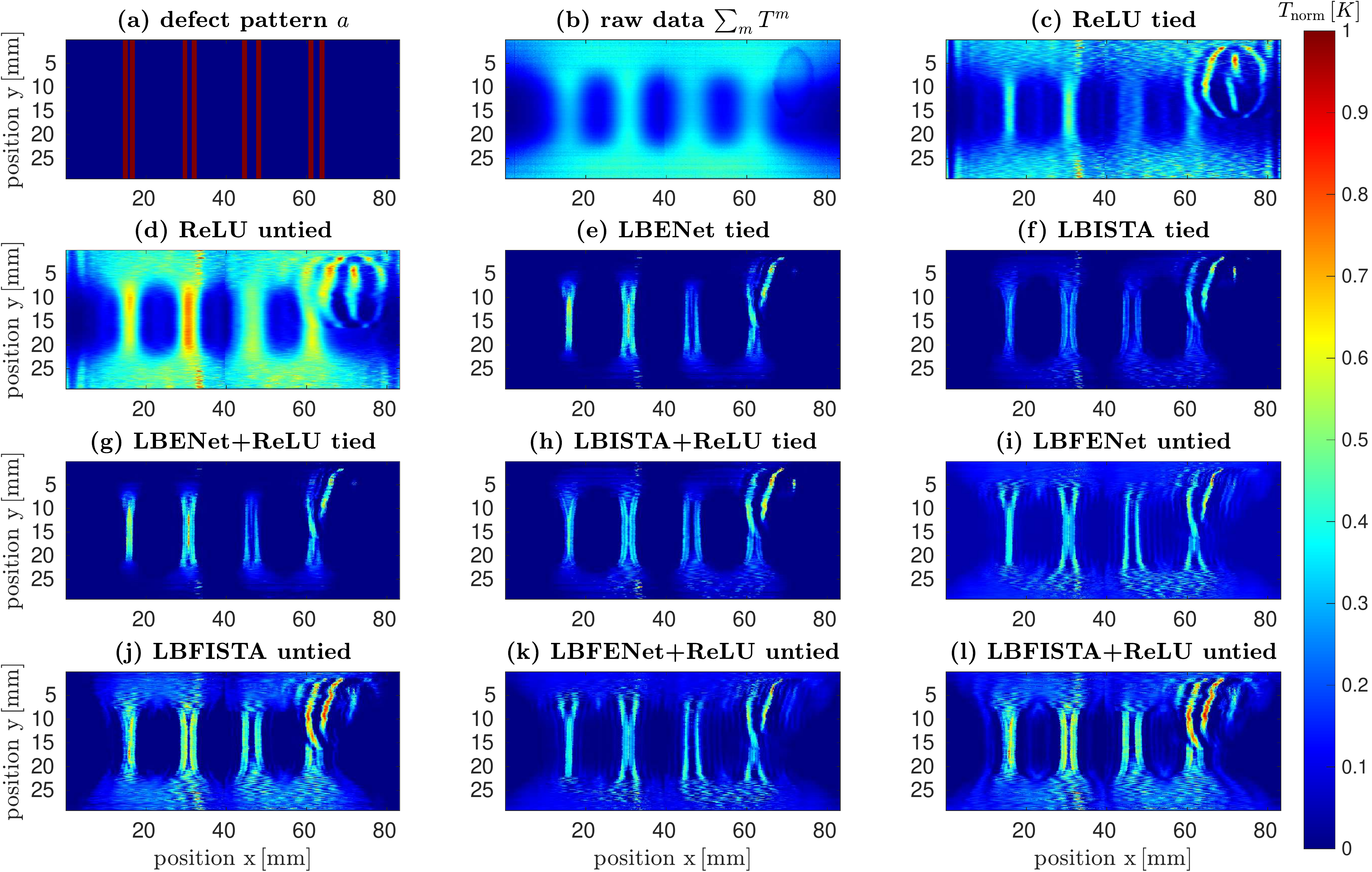}
    \caption{(a) real defect pattern as ideal result, (b) reconstruction based on raw data from structured laser illumination measurements, (c)-(l) reconstruction based on different deep unfolding approaches to realize photothermal SR applied to laser thermography data. An oval shaped damage can be seen in the upper right corner in the raw data (b), which influenced the reconstruction of the rightmost defect pair as shown in (c)-(l).}
    \label{fig:res_images}
\end{figure*}
Since with the proposed deep unfolding approach $u^m$ is reconstructed instead of the defect pattern $a$ (see Fig. \ref{fig:res_images} (a)), one can calculate the sum over all measurements and normalize this result. Normalization refers to a division by the highest amplitude value in the image. The normalized sum over all measurements of $u^m$ and the normalized $a$ are almost identical, as the laser scanned over the whole sample surface with small position shifts between the measurements (see detailed explanation in \cite{ahmadi2021multi} section 3 F). Calculating the sum over all measurements based on the raw data results in Fig. \ref{fig:res_images} (b). Only four defect regions can be observed instead of the actual eight defects. Moreover, these four defect regions can only be recognized in the region given by $y = 8 \dots 22\,$mm. The reason is that the used laser lines had a height of $10\,$mm so that only these $10\,$mm have been heated and the close surrounding area. In addition, there was an oval shaped damage on the sample surface, which can be seen in the upper right corner at around $x = 66 \dots 77\,$mm and $y = 2 \dots 15\,$mm. 

\noindent Applying the deep unfolding network only using ReLU with six layers, whether tied or untied, does not significantly improve the spatial resolution of defects as shown in Fig. \ref{fig:res_images} (c,d). In contrast, using deep unfolding with regularization in tied or untied learning, i.e. Learned Block-ENet, Learned Block-ISTA, Learned Block Fast ENet or Learned Block Fast ISTA (with or without ReLU) enables to resolve almost all defects with six layers accurately, as illustrated in Fig. \ref{fig:res_images} (e)-(l). 

\begin{figure*}[h]
    \centering
    \includegraphics[width = 0.75\textwidth]{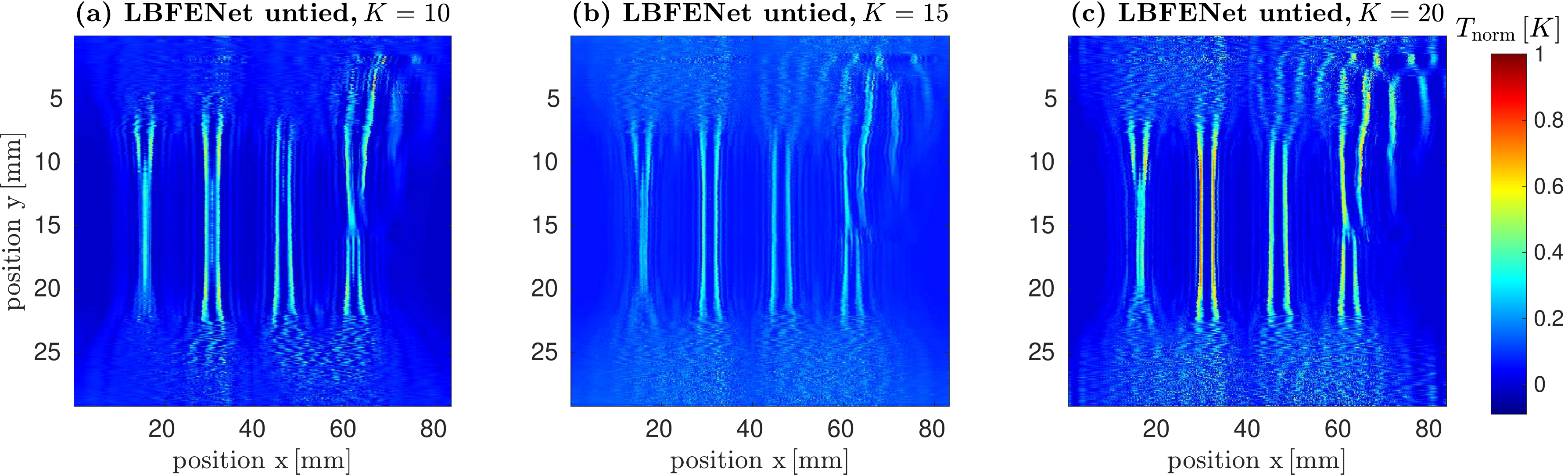}
    \caption{Untied LBENet with (a) 10, (b) 15 and (c) 20 layers.}
    \label{fig:layer_study}
\end{figure*}

\noindent It can be observed that the most challenging slit pair, i.e. the leftmost slit pair at around $x = 14 \dots 18\,$mm, with a distance between the slits of $0.5\,$mm, has never been reconstructed accurately. Furthermore, the damage in the upper right corner often precludes an accurate reconstruction of the rightmost slit pair. The other two slit pairs could be reconstructed precisely as shown for instance in Fig. \ref{fig:res_images} (j) and (l). The reconstruction quality is therefore largely determined by whether the defects can be separated from each other. The mean value of the laser illumination area is calculated, which is the region of interest, i.e. $y = 10 \dots 20\,$mm, and the resulting 1D vector is compared with the ideal defect pattern. This comparison can be formulated mathematically by a correlation coefficient. In the following, the Pearson correlation coefficient (see \textit{corrcoef}$(\,)$-function in Matlab) is used to determine the reconstruction quality quantitatively for all considered image processing methods. Table \ref{tab:rec_qualities} shows all calculated correlation coefficients.
\begin{table*}[h]
    \centering
    \begin{tabular}{|c||c|c|c|c|c|}  \hline
         \textbf{reconstruction quality}& \textbf{LBISTA}  & \textbf{LBFISTA} & \textbf{LBENet} & \textbf{LBFENet} & \textbf{no regularization} \\ \hline \hline 
         \textbf{tied} & 0.73 & 0.66 &  0.56 & 0.44 & - \\ \hline
         \textbf{untied} & 0.79 & 0.79 & 0.65 & 0.67 & -\\ \hline 
         \textbf{tied + ReLU} & 0.71 & 0.52 & 0.56 & 0.44 & 0.61\\ \hline 
         \textbf{untied + ReLU} & 0.78 & 0.79 & 0.65 & 0.69 & 0.61\\ \hline
    \end{tabular}
    \caption{Calculated reconstruction qualities based on Pearson correlation coefficient. The two vectors that have been used to calculate the correlation coefficient are computed by averaging over all measured pixels in $y = 10 \dots 20\,$mm in the final result image (some of them are shown in Fig. \ref{fig:res_images}).}
    \label{tab:rec_qualities}
\end{table*}
The following observations can be noted by evaluating Fig. \ref{fig:res_images} and table \ref{tab:rec_qualities}:

\begin{itemize}
    \item untied learning leads to a significant increase in reconstruction quality compared to tied learning
    \item the additional use of sparsity regularization based on LBISTA/ LBFISTA/ LBENet/ LBFENet improves the reconstruction quality, especially for untied learning
    \item the use of ReLU is not necessary as sufficiently high reconstruction accuracy is already reached by sparsity regularization in combination with untied learning
    \item reconstruction accuracy of up to $80\,$ \% can be achieved with Photothermal-SR-Net, whereas comparing Fig. \ref{fig:res_images} (a) and (b) leads to a reconstruction quality of only $50\,$\% and state-of-the art methods based on Block-FISTA and Block-ENet optimization without deep learning reach up to $60\,$\% \cite{ahmadi2020photothermal} which indicates that the manual choice of the regularization parameters in \cite{ahmadi2020photothermal} was not made perfectly
\end{itemize}
\textbf{More training, better performance.} \quad
An increase of the layer number can slightly improve reconstruction quality as illustrated in Fig. \ref{fig:layer_study}.
\begin{figure*}[h]
    \centering
    \includegraphics[width = 0.9\textwidth]{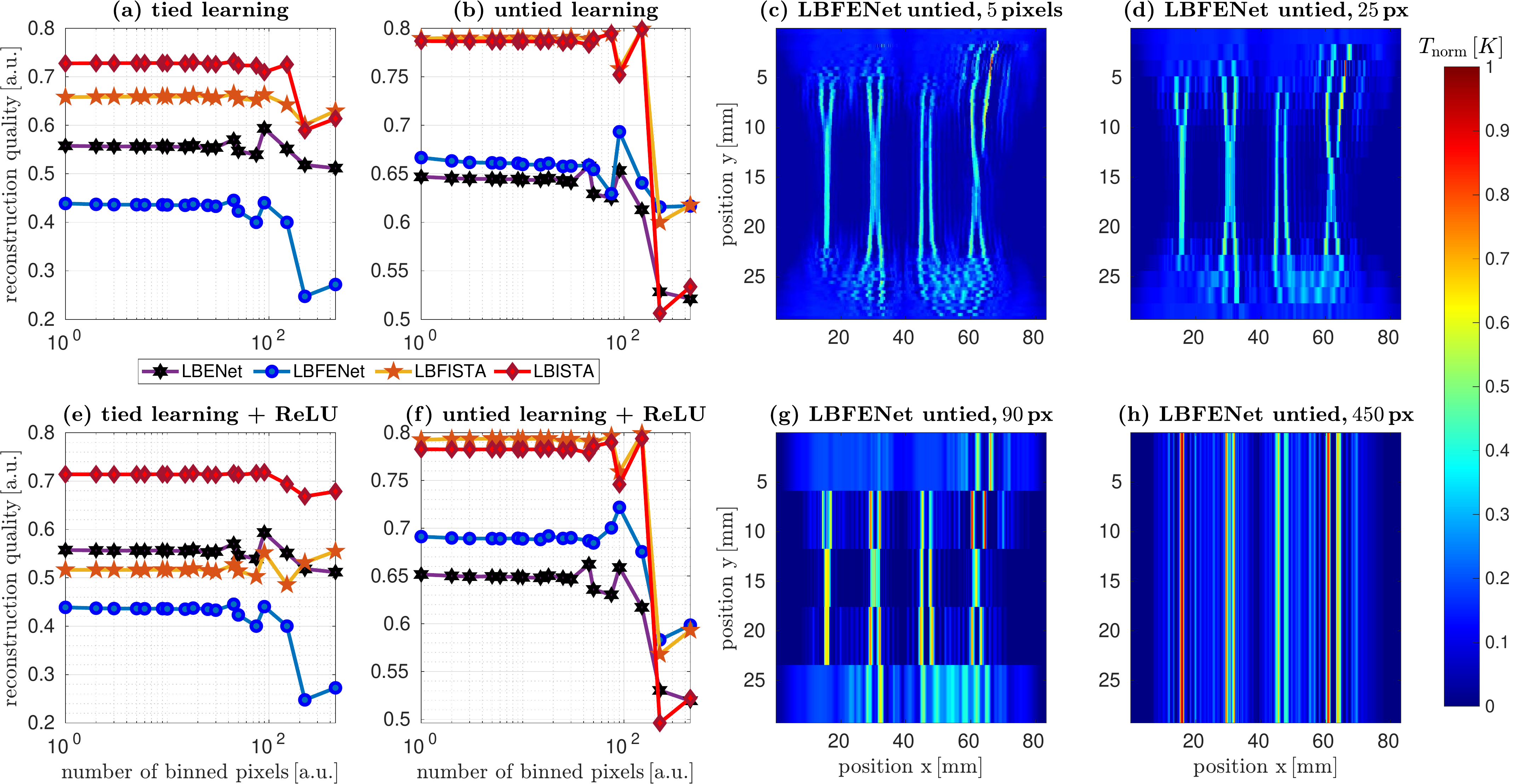}
    \caption{Influence of pixel binning, which was applied to test data, on final reconstruction quality after applying Photothermal-SR-Net to the binned test data.}
    \label{fig:px_binning}
\end{figure*} 
The corresponding reconstruction quality values for $K=10$, $K=15$ and $K=20$ are $0.69$, $0.67$ and $0.67$, respectively. According to the calculated values, a recognizable difference  in reconstruction quality is not observed. Nonetheless, a variation of the amplitude values of only some of the 1280 pixels in x-dimension could be crucial for the spatial resolution of two closely spaced defects. Fig. \ref{fig:layer_study} shows, for example, a more precise spatial resolution for the slit pair at around x-position $25$-$40\,$mm using $15$ (c.f. Fig. \ref{fig:layer_study} (b)) or even better with $20$ layers (c.f. Fig. \ref{fig:layer_study} (c)) instead of using 10 layers (c.f. Fig. \ref{fig:layer_study} (a)) or just 6 layers (c.f. Fig. \ref{fig:res_images} (i)).
\newline 
\textbf{Influence of pixel binning on reconstruction quality.} \quad
The idea is to employ pixel binning in the y-dimension, since the signal does not significantly change in the y-dimension, at least in the region of interest ($y = 10 \dots 20\,$mm). Pixel binning thus saves time and enables higher computational performance. In addition, it could lead to higher SNRs. On that account, the influence of pixel binning on the reconstruction quality in the final image was studied. Pixel binning was applied to the experimental raw data. Since $450\,$ pixels have been measured in the y-dimension, the studies were performed with the following numbers of binned pixels: $[1,\,2,\,3,\,5,\,6,\,9,\,10,\,15,\,18,\,25,\,30,\,45,\,50,\,75,\,90,\,150,\newline225,\,450]$, whereby $1$ means that no signal binning has been applied and equals to the results shown in Fig. \ref{fig:res_images}. These numbers have been chosen, as they represent all integer divisions of $450$. The influence on the reconstruction quality is shown in Fig. \ref{fig:px_binning} (a,b,e,f). The images in Fig. \ref{fig:px_binning} (c,d,g,h) illustrate how pixel binning changes the data to have a better imagination of the binning effect. LBFENet using untied learning was taken as an example to demonstrate the binning effect because the reconstruction quality could be improved by binning with $90$ pixels as one can observe in Fig. \ref{fig:px_binning} (b). Notably, pixel binning can be applied until around $30$ pixels without a significant change in the reconstruction quality. This saves computation time by a factor of $30$. A significant increase in the reconstruction quality cannot be realized by using pixel binning. In addition, the reconstruction quality clearly decreases if the chosen number of binned pixels is too high.
\newline 
\textbf{Computational performance of Photothermal-SR-Net.} \newline
The previous subsections have shown the level of reconstruction quality that can be obtained using Photothermal-SR-Net. Considering the underlying physics in the proposed deep unfolding approach, it was possible to achieve high convergence rates and to generate super resolved 2D thermal images using only six layers. With Photothermal-SR-Net, it takes around $100\,$ms to super resolve one pixel row of the thermal raw image. Since $450\,$ pixels have been investigated, it took around $45\,$s to generate a super resolved 2D thermal image of the entire sample surface. Using pixel binning as a preprocessing step allows to reduce the number of pixels. The previous subsection demonstrated that up to $30$ pixels could be binned without a decrease of reconstruction quality in the super resolved image. This indicates that only $15$ pixels instead of $450$ pixels have to be evaluated, resulting in a total time of merely $1.5\,$s to generate super resolved 2D thermal images.     
\section{Conclusion and Outlook}
This paper shows how super resolution using deep unfolding, namely Photothermal-SR-Net, can be applied to 2D thermal data based on synthetic 1D training. Since super resolution is required for the resolution of closely spaced defects in nondestructive testing, sparsity regularization with algorithms based on ISTA and Elastic-Net was mainly analyzed in terms of final reconstruction quality. In addition, it was shown that initial weighting based on the thermal point spread function could produce high-resolution reconstruction images with high convergence rates. It should be noted that updating the weights from layer to layer (untied instead of tied learning) is even more worthwhile, as more accurate reconstruction quality can be achieved. Furthermore, it could be shown that more layers lead to small but potentially relevant improvements in the reconstruction quality. It was also found that pixel binning as a preprocessing step can significantly increase the computational performance of Photothermal-SR-Net, enabling the generation of high-resolution images within one second without degrading the reconstruction quality. The ongoing work focuses on the adaption of the training loss to the underlying physics. In addition, a learning algorithm that selects the appropriate sparsity regularizer within deep unfolding will be implemented. Finally, it would be interesting to make use of block-sparsity not only over the number of measurements, but also in spatial dimensions. However, such an approach would require more knowledge about the defect structure.

{\small
\bibliographystyle{ieee_fullname}
\bibliography{egbib}
}

\end{document}